\documentclass[letterpaper, 10 pt, conference]{ieeeconf}  % Comment this line out if you need a4paper

\IEEEoverridecommandlockouts                              % This command is only needed if 
\usepackage[ruled]{algorithm2e} %,linesnumbered
\usepackage{setspace}
\usepackage{amssymb}
\usepackage{amsmath}
\usepackage{graphicx}
\usepackage{subfigure}

\title{\LARGE \bf
Exploring More When It Needs in Deep Reinforcement Learning}

\author{Youtian Guo and Qi Gao*% <-this % stops a space
\thanks{*This work was not supported by any organization}% <-this % stops a space
\thanks{The authors are with the Institute of Pattern Recognition and Intelligent Systems, Department of Automation, 
	Beijing Institute of Technology, Beijing, 100081, China 
	{\tt\small (e-mail: \{youtianguo, *corresponding author: Qi Gao\} @ bit.edu.cn)}}%
}

\begin{document}

\maketitle
\thispagestyle{empty}
\pagestyle{empty}

%%%%%%%%%%%%%%%%%%%%%%%%%%%%%%%%%%%%%%%%%%%%%%%%%%%%%%%%%%%%%%%%%%%%%%%%%%%%%%%%
\begin{abstract}

We propose a exploration mechanism of policy in Deep Reinforcement Learning, which is exploring more when agent needs, called Add Noise to Noise (AN2N). The core idea is: when the Deep Reinforcement Learning agent is in a state of poor performance in history, it needs to explore more. So we use cumulative rewards to evaluate which past states the agents have not performed well, and use cosine distance to measure whether the current state needs to be explored more. This method shows that the exploration mechanism of the agent's policy is conducive to efficient exploration. We combining the proposed exploration mechanism AN2N with  Deep Deterministic Policy Gradient (DDPG), Soft Actor-Critic (SAC) algorithms, and apply it to the field of continuous control tasks, such as halfCheetah, Hopper, and Swimmer, achieving considerable improvement in performance and convergence speed.

\end{abstract}

%%%%%%%%%%%%%%%%%%%%%%%%%%%%%%%%%%%%%%%%%%%%%%%%%%%%%%%%%%%%%%%%%%%%%%%%%%%%%%%%
\section{INTRODUCTION}

Policy exploration has always been one of the critical topic in the field of Reinforcement Learning (RL), agent's policy would diverge under excessive  exploration, However, if the exploration is not enough, policy is prone to converge prematurely. Part of exploratory research work focuses on the method of noise perturbations [1-2], or the method of entropy regularization [3,4], these methods are to explore the entire policy space, with strong randomness. The other part of the exploratory work is mainly to obtain a better policy by building a ``intrinsic'' reward [5-6].

In reinforcement learning tasks, epsilon-greedy is one of the most frequently applied exploration methods [1], but it does not carry out targeted exploration, so exponential data volume is required, as is Noisy Net [2]. Haarnoja et al. designed the Q value function into a boltzmann distribution form, which increases the diversity of policies [7]. Osband et al. [8] offered a promising approach to explore efficiently with generalization, which called randomized least-squares value iteration (RLSVI), but it is not suitable for non-linear value functions, such as neural networks. Osband et al. [9] developed bootstrapped Deep Q-Network (DQN), which combines deep exploration with deep neural networks. Subsequently, RLSVI was further extended to Multiplicative Normalizing Flows [10], which augments DQN and DDPG with multiplicative normalizing flows in order to track a rich approximate posterior distribution. 

Wealth of research is about how to design intrinsic rewards to help explore. Auer [11] proposed the confidence bounds method, which can be used to deal with situations which exhibit an exploitation-exploration trade-off in low-dimensional state space tasks. An extended of this work is that pseudo-count based method [12], which allocates rewards according to the  pseudo-count, and guide the agent to visit the state with a low count value. Yet this method is not applicable if the state space is high-dimensional. In order to improve the accuracy of pseudo-count, PixelCNN is proposed [13]. Zhao and Tresp applied Curiosity-Driven Prioritization (CDP) framework to encourage the agent to over-sample those trajectories that have rare achieved goal states [14], so as to develop the agent's exploration ability.In addition, there are many other extension work [15-16] related to Count-based exploration. Unlike cont-based, Houthooft et al. [17-19] use predictive models to adjust the intrinsic reward of the agent when exploring, Stadie et al. made use of an Auto Encoder (AE) to encode the state space, and estimated the agent's familiarity with the environment with deep predictive model [20], and then allocates rewards based on the predicted value of the model. Pathak et al. design exploration rewards based on disagreement of ensembles of dynamics models [21], which guides the agent to explore. For the purpose of alleviating the catastrophic forgetting of neural networks, Guo et al. used previously trained multiple policy models to interact with the environment to generate more training data for training the current policy network [22], so as to facilitate the agent to remember the explored state. 

In the past, the method of noise perturbation usually adds noise directly on policies, which requires a large amount of data interacting with the environment in high-dimensional action space tasks. Inspired by the Liebig's law of the minimum [23], we propose an Add Noise to Noise (AN2N) policy exploration method. The Liebig's law of the minimum shows that the capacity of a barrel with staves of unequal length is limited by the shortest stave, by analogy, we look uppon the policy improvement of the agent in RL as a process of building  or repairing a wooden barrel. For the sake of making the barrel hold more water at each step, we need to find the shortest stave and repair it higher. Similarly, in Reinforcement Learning, in order to help agents achieve better performance, we need to find the states that they need to explore most, and make the greater efforts to explore, which is the core idea of AN2N algorithm.

\section{Preliminaries}
Reinforcement learning considers the paradigm of an agent learning policies to maximize the expected reward in interacting with the environment. At each discrete time step $t$,  the agent receives an observation $o_t \in  \mathcal{O}$, selects actions $a_t \in \mathcal{A}$ with respect to its policy $\pi$: $\mathcal{O} \rightarrow \mathcal{A}$,  and receives a scalar reward $r_t$ and a next observation $o_{t+1}$ from the environment. In general, Reinforcement learning can be regarded as a Markov Decision Process (MDP) which  models stochastic, discrete-time and finite action space control problems [24-25]. A practical environment may always be partially observed, here, we assumed the environment is fully-observed, so $s_t=o_t, \mathcal{S}=\mathcal{O}$.

In reinforcement learning, the objective is to find the optimal policy $\pi$, which maximizes the expected return, a action-value function $Q^{\pi}$ is uesed to assess the quality of a policy $\pi$, defined as following:

$$
Q^{\pi}\left(s,a\right)=\mathbb{E}_{s_t \sim p_{\pi},a_t \sim \pi} \left[ \sum_{t=0}^{+\infty}\gamma^t R\left(s_t,a_t\right) \right] \eqno{(1)}
$$

Where $\gamma \in [0,1]$ is the discount factor determining the importance of future rewards, $\mathbb{E}_{s_t \sim p_{\pi},a_t \sim \pi}$ is the expectation return over the distribution of the trajectories $(s_0, a_0, s_1, a_1, \dots)$ obtained by performing action $a \sim \pi$ in state $s \sim p_{\pi}$.

The action-value function of the optimal policy is the largest, which is $Q^*(s, a) = \mathop{\arg\max}_{\pi} Q^{\pi}(s, a)$, the value function $V^\pi$ is the mean value of $Q^{\pi}$ obtained by selecting action $a$ according to policy $\pi(\cdot|s)$ distribution in state $s$, defined as $V^\pi(s)=\mathbb{E}_{a \sim \pi(\cdot|s)} \left[Q^{\pi}(s, a)\right]$. Since we consider reinforcement learning as an MDP problem, we can express action-value function $Q^{\pi}$ in the form of dynamic programming:

$$
\begin{aligned}
	Q^{\pi}\left(s_t,a_t\right) &=\mathbb{E}_{s_{t+1} \sim p_{\pi}} [r(s_t,a_t) \\
	&+\gamma \mathbb{E}_{a_{t+1} \sim \pi} \left[Q^{\pi}(s_{t+1}, a_{t+1})\right]]
\end{aligned}
\eqno{(2)}
$$

In low dimensional state-action space tasks, the $Q^{\pi}$ function in (2) is usually expressed as look-up table method, for example in Q-Learning [26]. In pace with the dimension of state-action space becomes higher, the look-up table method is becoming less and less applicable, expecially in complex tasks. Therefore, Deep Reinforcement Learning (DRL) uses deep neural networks as function approximators for RL methods [27], Then, more and more algorithms are proposed in the field of deep reinforcement learning, such as Deep Deterministic Policy Gradient(DDPG) [28], Trust Region Policy Optimization [29], Asynchronous Advantage Actor-Critic (A3C) [30], Soft Actor-Critic(SAC) [4] and Twin Delayed Deep Deterministic Policy Gradient(TD3) [31] algorithms. 

DDPG applied neural network to approximate the action-value function $Q(s,a|\theta^Q)$ and policy function $\mu(s|\theta^\mu)$, called  critic network and actor network, respectively, with the parameters $\theta^Q$, $\theta^\mu$,  the DDPG algorithm introduces critic target network $\theta^{Q^{'}}$ and policy target network $\theta^{\mu^{'}}$, so as to improve the stability of policy update. Consequently, gradient descent is used to optimize the network weight by minimizing the loss:

$$
\begin{aligned}
	L(\theta^Q)=\mathbb{E}_{s_t \sim p_{\mu(s_t|\theta^\mu)},a_t \sim \mu(s_t|\theta^\mu)} &[(Q(s_t,a_t|\theta^Q) \\
	&-y_t)^2] 
\end{aligned}
\eqno{(3)}
$$
Where
$$
y_t=r(s_t,a_t)+\gamma Q^{'}(s_{t+1},\mu^{'}(s_{t+1}|\theta^{\mu^{'}})|\theta^{Q^{'}}) \eqno{(4)}
$$
$$
\begin{aligned}
	\nabla_{\theta^{\mu}}J &\approx  \mathbb{E}_{s \sim p_{(s_t|\theta^{\mu})}}\left[ \nabla_{\theta^{\mu}} Q(s,a|\theta^Q ) |_{s=s_t,a=\mu(s_t|\theta^{\mu})}\right] \\
	&= \mathbb{E}_{s \sim p_{(s_t|\theta^\mu)}} [ \nabla_a Q(s,a|\theta^Q) |_{s=s_t,a=\mu(s_t)} \\
	&\qquad \qquad \qquad \quad \nabla_{\theta^{\mu}}\mu(s_t|\theta^\mu)|s=s_t ] 
\end{aligned}
\eqno{(5)}
$$

Equation (4) derived from (2), the target actor network decouples the process of policy updating and policy improving,  and the weights of critic and policy target network are either updated periodically to slowly track the learned networks: $\theta^{'}\leftarrow \tau \theta +(1-\tau)\theta^{'}$ with $\tau \ll 1$, which avoids the large fluctuation in the agent's learning process.  The actor is updated by (5), following the chain rule to the expected return $Q(s,a|\theta^Q )$ from the distribution $J$ with respect to the actor parameters $\theta^{\mu}$.

\section{Exploring More When It Needs}
In reinforcement learning environment, agent often selects different action in different state. Due to the vulnerability of the policy, it is presumable for agent to perform terribly in some states, agent proceed to the next step,  getting into a new state that has not been learned before. The terrible policy begin to affect the decisions of the following states, and ultimately affect the overall performance of the agent. We decompose this problem into three sub problems:
\begin{itemize}
	 \item When the agent needs to explore as much as possible? 
	 \item How to determine whether the current state needs to explore more?
	 \item How to explore? 
\end{itemize}
The solution of these three problems is also the core idea of our proposed AN2N algorithm.
\subsection{Exploring More When Agent in a Bad State} 
As the above analysis shows, due to the vulnerability of the policies, the agent may be in a dilemma in some states. Once an agent falls into a terrible state, it is likely to have an impact on the following trajectory, thus affecting the overall performance. This reminds us of the Liebig's law of the minimum, which indicates that the capacity of a barrel is limited by the length of the shortest stave. As shown in Fig. 1, for the sake of effectively improving the capacity of the barrel, it is necessary to lengthen the shortest stave first. See more details in Fig. 5 in Appendix.

\begin{figure}[thpb]
	\centering
	\includegraphics[scale=0.4]{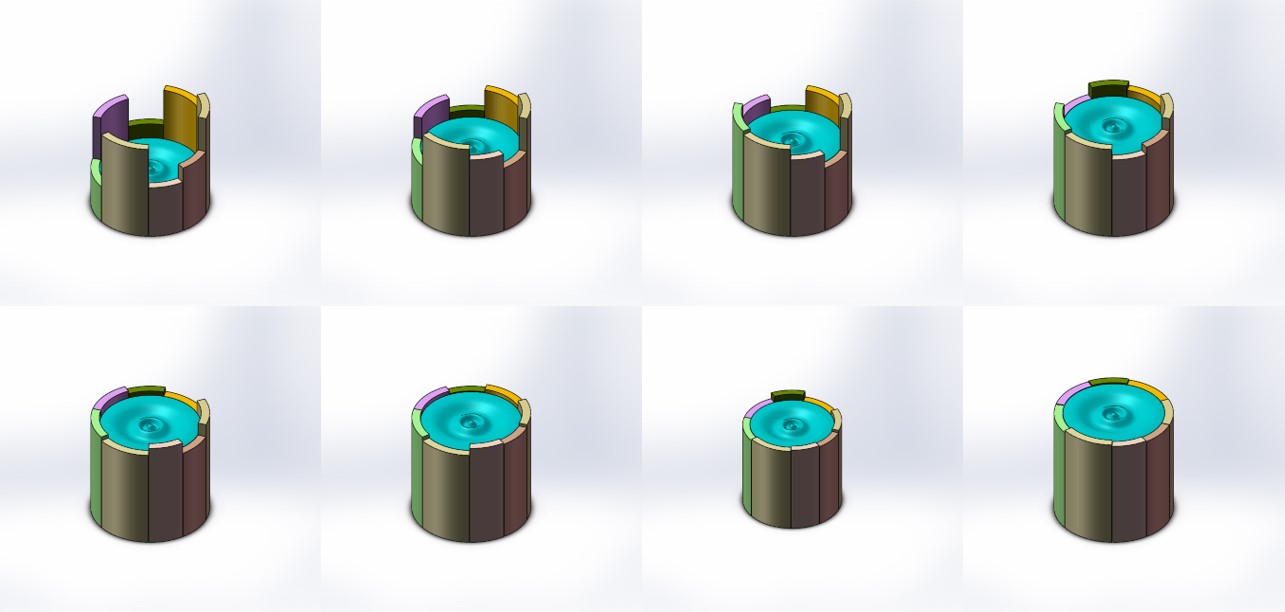} %[????]{????}
	\caption{Increasing the capacity of a barrel according to the Liebig's law of the minimum}
	\label{fig: figure1}
\end{figure}

Similar to the principle of repairing short stave, agents need to focus on these poor performance states and explore more to stabilize the process of policy improvement. So we use the cumulative reward of the state to evaluate whether the state is bad, equation (4) provides a solution, but it's a one-step Q-learning which obtains a reward $r$ and only directly affects the value of the state action pair $s, a$, The other state action pairs are affected indirectly by updating $Q(s,a)$ function, which slowdown the learning process since many updates required to propagate a reward to the relevant preceding states and actions. Hence, we choose n-step returns [26, 32] that propagates rewards faster, defined as:

$$
\begin{aligned}
	Reward(s_t) &=r_t+\gamma r_{t+1}+\cdots	+\gamma^{n-1}r_{t+n-1}\\ &+\max_{a}\gamma^{n}Q(s_{t+n},a) 
\end{aligned}
\eqno{(6)}
$$

Where current state reward affecting the values of n preceding state action pairs directly, which makes the process of propagating rewards to relevant state-action pairs potentially much more efficient. For the purpose of applying it to our algorithm, we rewrite it as follows:
$$
\begin{aligned}
	Reward(s_t) &\approx r_t+\gamma r_{t+1}+\cdots+\gamma^{T-t+1}r_{T-1}\\
	&+\gamma^{T-t}Q^{'}(s_T,\mu(s_T|\theta^{\mu})|\theta^{Q^{'}}) 
\end{aligned}
\eqno{(7)}
$$

We can calculate the reward value of each state according to (7) after the agent generates a trajectory, and greedily select the worst state to store in the fixed length FIFO queue.

\subsection{Calculate the Similarity between States}

When the agent interacts with the environment, it is necessary to determine whether the current interaction state needs to be explored. We use similarity measurement to judge if the current state is similar to the state in FIFO queue. The current state needs to be explored more if it is similar.

Similarity Mesurement is widely used in the field of Recommender Systems, we select two kinds of distance to measure the similarity between different states, namely Manhattan distance and Cosine distance:
$$
\begin{aligned}
	Manhattan\_sim(s_i,s_j) = \frac{1}{1+\sum_{k}\left| s_{i,k}-s_{j,k} \right|} 
\end{aligned}
\eqno{(8)}
$$
$$
\begin{aligned}
	&Cosine\_sim(s_i,s_j) = \\
	&\frac{\sum_{k}(s_{i,k}-\bar{s_k}) \cdot (s_{j,k}-\bar{s_k})}{\sqrt{\sum_K(s_{i,k}-\bar{s_k})^2}\cdot\sqrt{\sum_K(s_{j,k}-\bar{s_k})^2}}
\end{aligned}
\eqno{(9)}
$$

Equation (8) describes Manhattan distance similarity, which is able to capture local differences between states, while cosine distance similarity in (9) measures the difference as a whole. Since these method needs a threshold to judge whether two states are similar or not, we set an decayed variable $Pct_{add}$, means the proportion of bad states in the total interaction state, $Pct_{add}$ was used to automatically adjust the similarity threshold, if it'is too high, the similarity threshold will be increased, otherwise, decreased.

\subsection{Add Noise to Noise}

Agent knows whether the current state needs more exploration under the similarity of the bad states, those who need to be explored more called key states. A lot of exploration methods are analyzed in Section 1, we choose one of the most simple and effective methods to verify our method AN2N, that is, adding noise perturbations to the policy. When the agent interacts with the environment normally,  it needs to add a small noise disturbance to the policy $\mu(s_t|\theta^\mu)$, so as to ensure the basic exploration ability of the policy. When the agent is in the key states, it needs to add a noise to the small noise, or directly add a big noise $\mathcal{N}_{\textrm{big}}=\mathcal{N}_{\textrm{noise}}+\mathcal{N}_{\textrm{noise}_{\textrm{add}}}$ to increase the exploration, which is also the origin of the algorithm name (Add Noise to Noise, AN2N). The pseudo code of AN2N algorithm is shown in algorithm~\ref{AN2N}.

\begin{algorithm}[h]
	\begin{spacing}{0.8}
		\caption{AN2N} \label{AN2N}
		\KwIn{Noise  $\mathcal{N}_{\textrm{small}}$,  $\mathcal{N}_{\textrm{big}}$, Replay buffer size  $R$, $R_{\textrm{AN2N}}$, $FIFO_{\textrm{AN2N}}$,  Key states $K_{\textrm{upper}}$,  $K_{\textrm{lower}}$}
		Randomly initialize critic network $Q(s,a|\theta^Q)$ and actor $\mu(s|\theta^\mu)$ with weights $\theta^Q$ and $\theta^\mu$\\
		Initialize target network $Q^{'}$ and $\mu^{'}$ with weights $\theta^{Q^{'}} \leftarrow  \theta^Q$, $\theta^{\mu^{'}} \leftarrow  \theta^{\mu}$\\
		
		%\KwOut{Observation $S_0$ and choose$A_0\sim\pi_\theta(S_0$)}
		
		\For{episode $ e \in \{$1,...,M$\}$}
		{Initialize a random process $\mathcal{N}$ for action exploration\\
			Receive initial observation state $s_1$\\
			\For{$t \in \{$1,...,T$\}$}
			{
				\# $S$ is the set of states in $FIFO_{AN2N}$\\
				\uIf {\rm{Similarity($s_t, S$)}} {
					$\mathcal{N}_t$ = $\mathcal{N}_{\textrm{big}}$
				} \Else{
					$\mathcal{N}_t$ = $\mathcal{N}_{\textrm{small}}$
				}
				Select action $a_t=\mu(s_t|\theta^\mu)+\mathcal{N}_t$ 
				according to the current policy and exploration noise\\
				Execute action $a_t$ and observe reward $r_t$ and observe new state $s_{t+1}$\\
				Store transition $(s_t, a_t, r_t, s_{t+1})$ in $R$\\
				Test the agent and store the trajectory $(s_t, r_t)$ in $R_{AN2N}$\\
				Calculate the cumulative discount rewards of each state:\\
				\begin{center}
					\vspace{2ex}
					$Reward(s_t) \approx r_t+\gamma r_{t+1}+\cdots+\gamma^{T-t+1}r_{T-1}+\gamma^{T-t})Q^{'}(s_T,\mu(s_T|\theta^{\mu})|\theta^{Q^{'}})$
				\end{center}
				Save the clip$({20\times(\frac{\textrm{average} \  \textrm{reward}}{\textrm{reward}}})^2, K_{\textrm{lower}}, K_{\textrm{upper}}) $\\  $Reward(s)$ minimum key states in $FIFO_{\textrm{AN2N}}$\\
				Run DDPG, SAC or TD3 etc. Algorithms
			}
		}
	\end{spacing}
\end{algorithm}

\section{Result}

We choose two representative algorithms to combine with AN2N. The first one is DDPG, which uses neural network to represent action policy $\mu(s_t|\theta^\mu)$ for the first time, thus extending the application of deep reinforcement learning from discrete control to continuous control. It is one of the most famous algorithms in the field of continuous control. The second one is SAC, an off policy algorithm based on maximizing policy entropy, it is still a state of the art algorithm benefit from it's good exploration, and has a good landing application in the industry.

\begin{figure}[thpb]
	\begin{center}
		\includegraphics[width=0.45\textwidth]{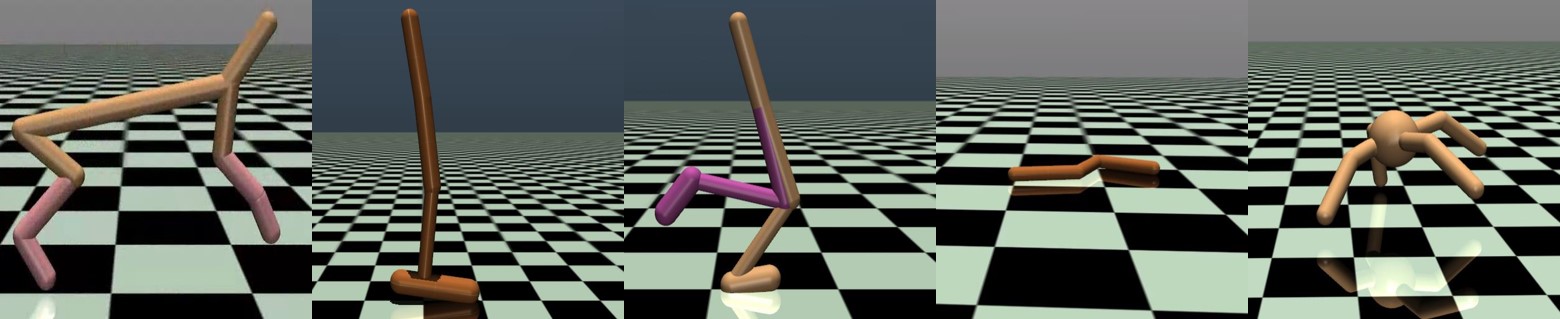}%spiral.eps
		\caption{Samples of environments we attempt to solve with AN2N. In order from the left: make a 2D cheetah robot run, make a two-dimensional one-legged robot hop forward as fast as possible, make a two-dimensional bipedal robot walk forward as fast as possible, make a 3-link swimming robot swim forward as fast as possible in a viscous fluid, make a four-legged creature walk forward as fast as possible}\label{fig:figure2}
	\end{center}
\end{figure}

We evaluate our algorithm combined with DDPG and SAC on 5 continuous control tasks of varying levels of difficulty, all of which are simulated using the MuJoCo physics engine [34], as it offers a unique combination of speed, accuracy and modeling power, and it is the first full-featured simulator designed from the ground up for the purpose of motion control, illustrated in Fig. 2. To test the generalization of the algorithm, we kept the same hyperparameters in different environments.

As introduced in Section 2, DDPG uses two actor-networks (acotr-network: $\mu(s|\theta^\mu)$ and target acotr-network:$\mu^{'}(s|\theta^{\mu^{'}})$) and critic-networks (critic-network: $Q(s,a|\theta^Q)$ and target critic-network:$Q^{'}(s,a|\theta^{Q^{'}})$) respectively to approximate the policy and action-state value.  When the agent interacts with the environment, it first uses random policy to obtain some interaction data for the initial training of the networks, and then starts to use the policies of superimposing disturbance noise to interact with the environment. In the test phase, it records the reward of each state of the agent,  and calculates the action state value of the last state according to (7), the pseudo code of DDPG with AN2N is shown in algorithm~\ref{algorithm1} in Appendix B. It should be noted that in AN2N, the superimposed small noise value is set to 0.05, the large noise value is set to 0.4, and the proportion of large noise $Pct_{add}$ linearly decays from 0.4 to 0.2, which limits the noise integral value in the whole interaction process to a reasonable range.

\begin{figure}[htbp] %htbp
	\centering
	\subfigure[HalfCheetah]{\includegraphics[width=0.15\textwidth]{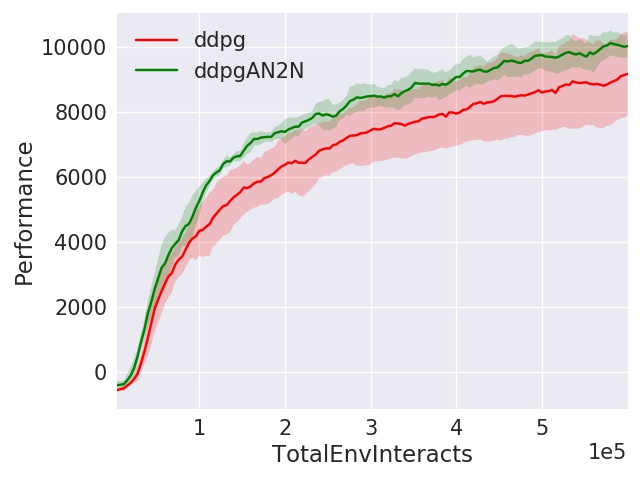}} 
	\subfigure[Hopper]{\includegraphics[width=0.15\textwidth]{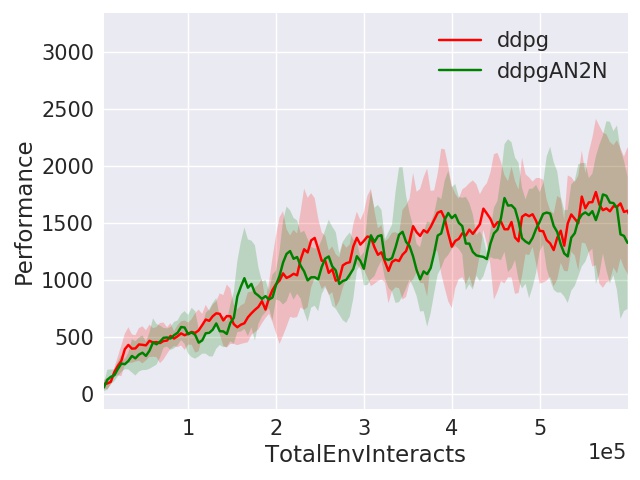}}	
	\subfigure[Walker2d]{\includegraphics[width=0.15\textwidth]{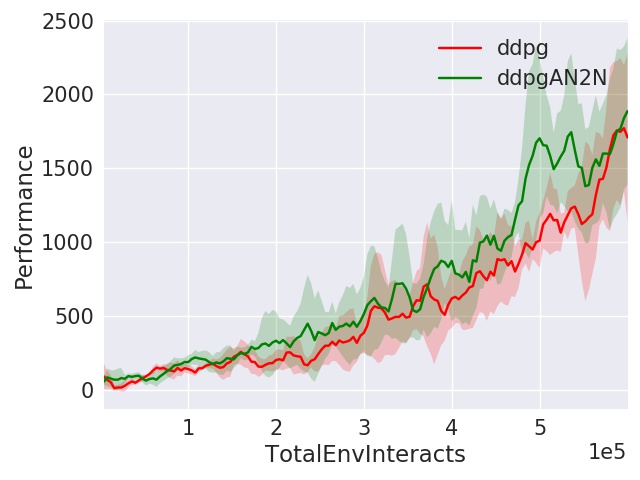}}
	\\ %??
	\centering
	\subfigure[Swimmer]{\includegraphics[width=0.23\textwidth]{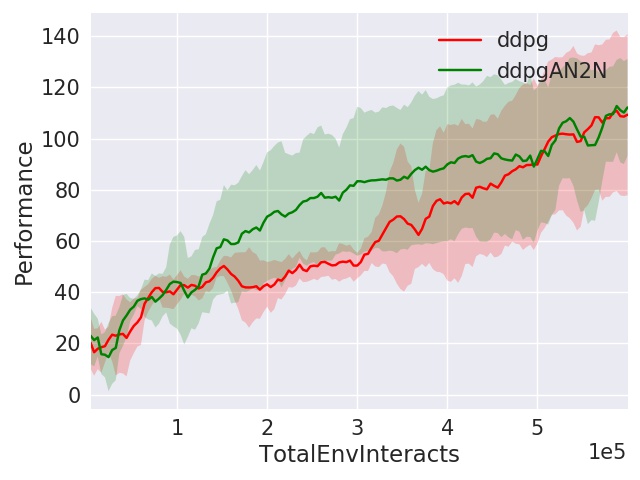}}
	\subfigure[Ant]{\includegraphics[width=0.23\textwidth]{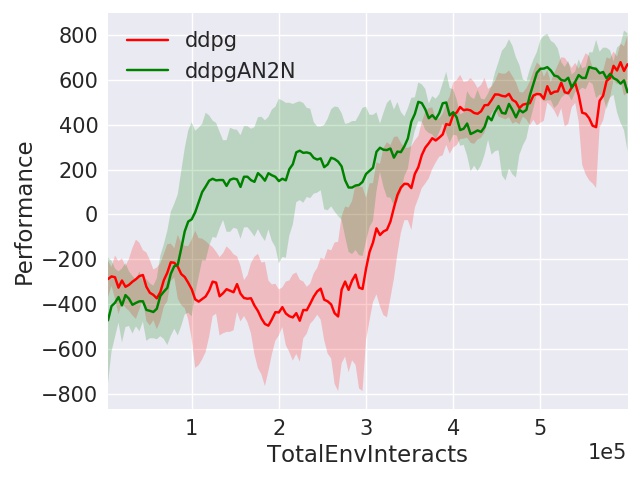}} 
	%\subfigure[HalfCheetah]{\includegraphics[width=0.15\textwidth]{paperpic/ddpgsac_HalfCheetah.jpeg}} 
	\caption{DDPG agent combined with AN2N versus the baseline Within 6e5 timestep for different test environments. (a) ddpgAN2N agent versus the baseline in HalfCheetah-v2. (b) ddpgAN2N agent versus the baseline in Hopper-v2. (c) ddpgAN2N agent versus the baseline in Walker2d-v2. (d) ddpgAN2N agent versus the baseline in Swimmer-v2. (e) ddpgAN2N agent versus the baseline in Ant-v2.} %????
	\label{fig:figure3}  %??????????
\end{figure}

%\twocolumn 
\begin{table*}[h]
	\caption{Mean value of cumulative reward of agent in different test environments}
	\label{tab:table3}
	\centering
	\begin{tabular}{c | c c c c c}\hline
		\textbf{Environment}       &\textbf{Random}        &\textbf{DDPG}            & \textbf{DDPG with AN2N}  & \textbf{SAC}              & \textbf{SAC with AN2N}            \\\hline
		HalfCheetah & -284$\pm$27  & 6550 $\pm$ 1291 & 7541 $\pm$ 651  & 8326 $\pm$ 1577  & $\mathbf{8803 \pm 579}$ \\
		Hopper      & 18$\pm$6     & 1659 $\pm$ 992  & 1067  $\pm$ 726 & 2348 $\pm$ 637   & $\mathbf{2562 \pm 573}$   \\
		Walker2d    & 2$\pm$2      & 541 $\pm$ 361   & 685 $\pm$ 455   & 2566 $\pm$ 765   & $\mathbf{2834 \pm 752}$  \\
		Swimmer     & 0$\pm$4      & 63 $\pm$ 26     & $\mathbf{74 \pm 31}$  & 41 $\pm$ 2       & 41 $\pm$ 2                 \\
		Ant         & -58$\pm$35   & 24 $\pm$ 319    & 245 $\pm$ 342   & 1595 $\pm$ 848   & $\mathbf{1835 \pm 989}$ \\\hline
	\end{tabular}
\end{table*}
The training process of Deep Reinforcement Learning often fluctuates a lot as the instability of policy update and policy improvement, in order to increase the credibility of the experiment result, each simulation environment simulates $6\times 10^5$ steps, and every 4000 steps is set as an epoch, where the learned policies are tested 10 times, and the average value is taken as the performance of the test, and we use the same set of parameters in the five environments in Fig. 2, and repeat the experiment with five different random seeds in each environment, seeds are set to 0, 5, 10, 15 and 20 respectively.

%\onecolumn 
%\begin{table*}[!htbp]
%	\caption{MaxTestEpRet}
%	\label{tab:table2}
%	\centering
%	\begin{tabular}{c | c c c c c}\hline
%		Games       &Random & DDPG            & DDPG with AN2N         & SAC              & SAC with AN2N           \\\hline
%		HalfCheetah & -38   & 9770 $\pm$ 1926 & 10787 $\pm$ 658        & 10862 $\pm$ 3504 & $\mathbf{11730 \pm 319}$\\
%		Hopper      & 169   & $\mathbf{3872 \pm 45}$  & 3494 $\pm$ 97  & 3523 $\pm$ 213   & 3587 $\pm$ 124          \\
%		Walker2d    & 44    & 4753 $\pm$ 979  & 4465 $\pm$ 751         & 4609 $\pm$ 388   & $\mathbf{4947 \pm 539}$ \\
%		Swimmer     & 43    & 148 $\pm$ 30    & $\mathbf{181 \pm 71}$  & 50 $\pm$ 7       & 49 $\pm$ 5              \\
%		Ant         & 67    & 1533 $\pm$ 546  & 1800 $\pm$ 433         & 4188 $\pm$ 1547  & $\mathbf{4472 \pm 2534}$\\\hline
%	\end{tabular}
%\end{table*}

\begin{figure}[htbp] %htbp
	\centering
	\subfigure[HalfCheetah]{\includegraphics[width=0.15\textwidth]{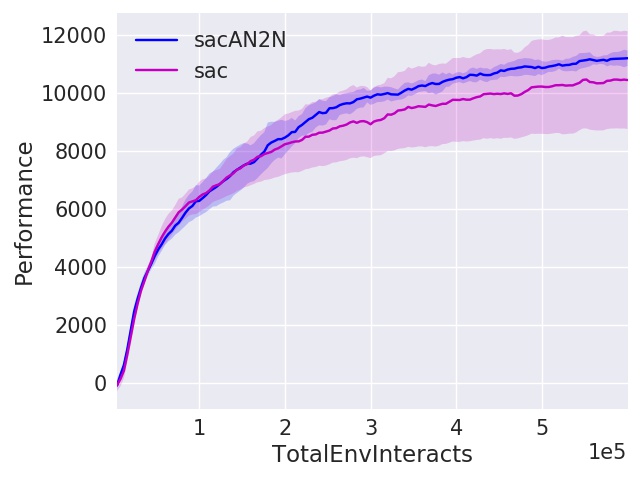}} 
	\subfigure[Hopper]{\includegraphics[width=0.15\textwidth]{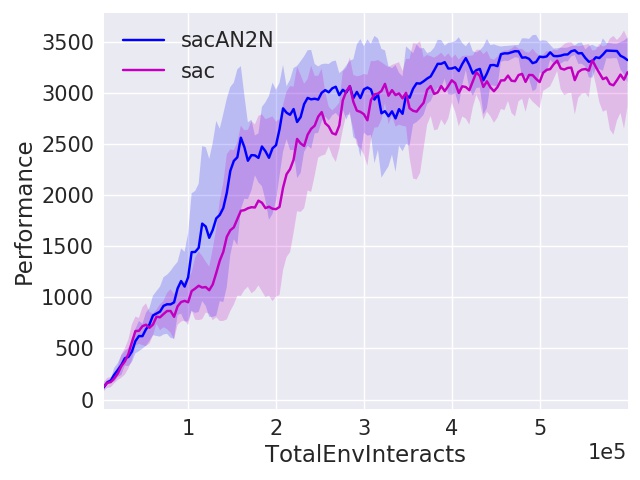}}	
	\subfigure[Walker2d]{\includegraphics[width=0.15\textwidth]{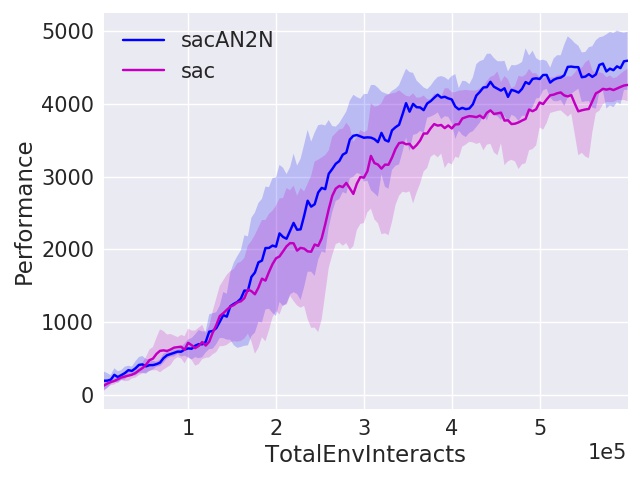}}
	\\ %??
	\centering
	\subfigure[Swimmer]{\includegraphics[width=0.23\textwidth]{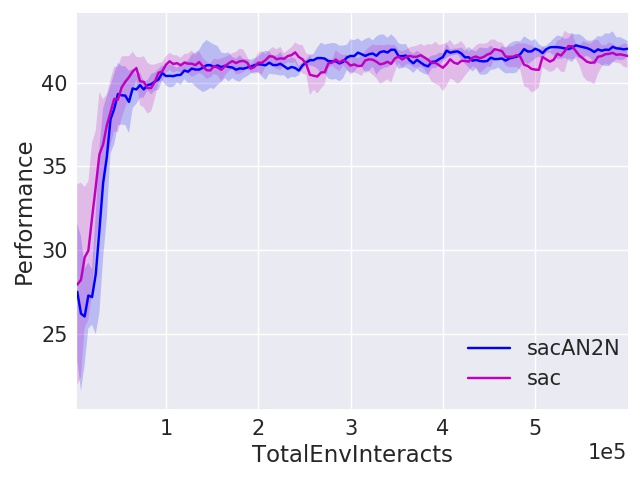}}
	\subfigure[Ant]{\includegraphics[width=0.23\textwidth]{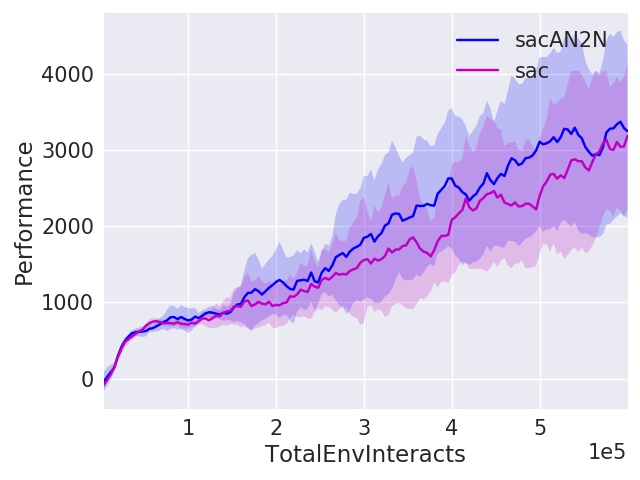}} 
	\caption{SAC agent combined with AN2N versus the baseline Within 6e5 timestep for different test environments.} %????
	\label{apdfig1}  %??????????
\end{figure} 

The experimental results of DDPG with AN2N and DDPG benchmark are shown in Fig.~\ref{fig:figure3}, our algorithm achieves better performance in tasks HalfCheetah and Walker2d, and faster convergence speed in tasks HalfCheetah, Walker2d, Swimmer and Ant. 

Similarly, we combine SAC with AN2N. The Q function is usually updated by updating the bellman residual, while the SAC add the policy entropy term, and the output of SAC policy network is a distribution, which is generally expressed by the mean and variance of Gaussian distribution. Since Benchmark limits the variance of policy output, in order to combine SAC and AN2N more succinctly, we increase the variance limit range by $1.5$ times in the key states where we need to explore more, and the variance is reduced to $0.5$ times the original in other states. The pseudo code of SAC with AN2N is shown in algorithm \ref{algorithm2} in Appendix. Compared with the Benchmark of SAC in the test environment in Fig. 2, the experimental results are shown in Fig. 4. Though SAC has higher stability and better performance than DDPG and other algorithms, SAC with AN2N has better performance in convergence speed and performance in continuous action control tasks such as HalfCheetah, Hopper Walker2d etc..

We summarize and present the experimental results in Table 1, and the random means agent taking a randomly generated policy. Each value represents the average return over 10 trials of 0.6 million time steps in five different seeds, the maximum value for each task is bolded. $\pm$ corresponds to a single deviation over trials. *AN2N matches or outperforms all baselines in both final performance and learning speed across all tasks, especially SAC combined with AN2N, which shows thatthe combination of AN2N with DDPG and SAC achieve a significant performance improvement effect.

%
%Finally, we draw the experimental results of DDPG with AN2N and SAC with AN2N on the same graph for comparison. The results are shown in the subgraph $f$ in Fig. 3 . SAC has made more improvements than DDPG in terms of the update process of policy network, the calculation of action-value function, and the generation method of action. Therefore, the performance of SAC is obviously better than that of DDPG, but after the combination of DDPG and AN2N, The performance of DDPG with AN2N method is close to that of SAC algorithm, which shows the simplicity and efficiency of AN2N Algorithm for convergence speed and performance.

%
%$$
%\begin{aligned}
%	100\times \frac{\textrm{Score}_{\textrm{AN2N}}-\textrm{Score}_{\textrm{Baseline}}}{\textrm{Score}_{\textrm{Baseline}}-\textrm{Score}_{\textrm{Random}}}  \label{Improvement}
%\end{aligned}
%\eqno{(10)}
%$$
%
%\begin{table}[h]
%	\centering
%	\caption{Margin settings for A4 size paper}
%	\label{tab:table1}
%	\begin{tabular}{c | c c c}\hline
%		Mean&Baseline & AN2N & Improvement\\\hline
%		DDPG & $\mathbf{ 355 }$ &  25 &  15\%     \\
%		SAC & $\mathbf{ 655 }$ &$\mathbf{ 155 }$ &25 \\\hline
%	\end{tabular}
%\end{table}

\section{Conclusion}

We propose a novel policy exploration method which called AN2N based on the Liebig's law of the minimum, owing to its excellent scalability,  AN2N can be well combined with the currently frequently used  algorithms such as DDPG, SAC, which enhances its exploration ability. AN2N algorithm is divided into the following three steps: 1. Use the idea of n-step Q-learning to calculate the return of each state,used for measuring which states are prone to get the agent into a dilemma, and preserve them; 2. Compare the current state with the dilemma state, if similar, the current state needs to explore more, and make use of the proportion of added noise to automatically adjust the similarity threshold. 3. Add noise on the noise to increase the intensity of exploration. We combine AN2N with DDPG and SAC algorithms to verify its performance in the mainstream test environments of continuous control tasks, and achieve significant improvement in performance and convergence speed.

\addtolength{\textheight}{-3.5cm}   % This command serves to balance the column lengths

%%%%%%%%%%%%%%%%%%%%%%%%%%%%%%%%%%%%%%%%%%%%%%%%%%%%%%%%%%%%%%%%%%%%%%%%%%%%%%%%

%%%%%%%%%%%%%%%%%%%%%%%%%%%%%%%%%%%%%%%%%%%%%%%%%%%%%%%%%%%%%%%%%%%%%%%%%%%%%%%%

%%%%%%%%%%%%%%%%%%%%%%%%%%%%%%%%%%%%%%%%%%%%%%%%%%%%%%%%%%%%%%%%%%%%%%%%%%%%%%%%
\section*{APPENDIX}
Appendix includes: the schematic diagram of the whole process of increasing the capacity of a barrel in Fig. 5, pseudo code of DDPG with AN2N in \textbf{Algorithm}~\ref{algorithm1},  pseudo code of SAC with AN2N in \textbf{Algorithm}~\ref{algorithm2}.
\begin{figure}[h]
	\centering
	\includegraphics[width=0.35\textwidth]{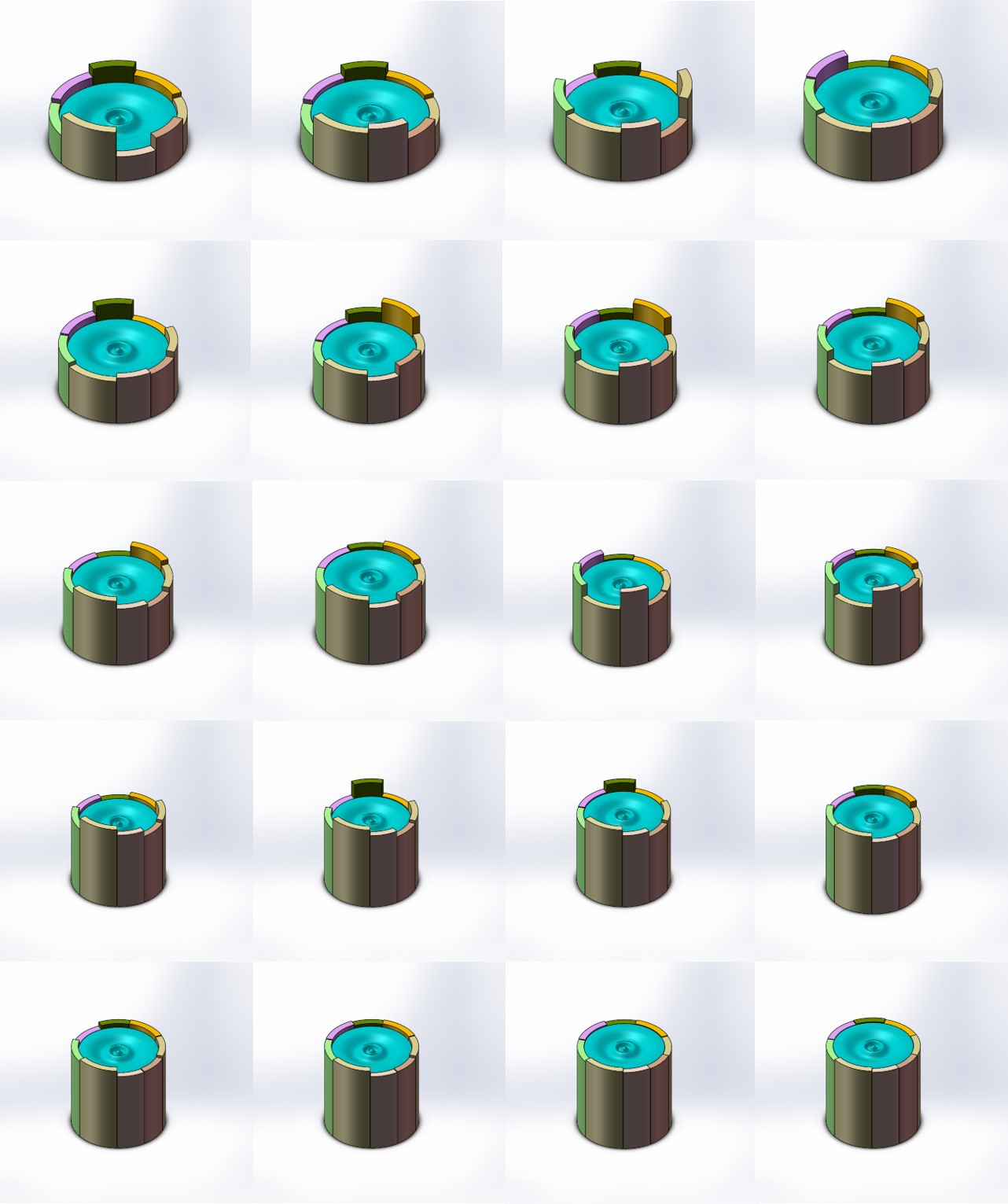} %[????]{????}
	\caption{The whole process of increasing the capacity of a barrel according to the Liebig's law of the minimum} %????
	\label{apdfig: figure7}
\end{figure}

\begin{algorithm}[h]
	\begin{spacing}{0.8}
		\caption{DDPG with AN2N}\label{algorithm1}
		\KwIn{Noise  $\mathcal{N}_{\textrm{small}}$,  $\mathcal{N}_{\textrm{big}}$, Replay buffer size  $R$, $R_{\textrm{AN2N}}$, $FIFO_{\textrm{AN2N}}$,  Key states $K_{\textrm{upper}}$,  $K_{\textrm{lower}}$}
		Randomly initialize critic network $Q(s,a|\theta^Q)$ and actor $\mu(s|\theta^\mu)$ with weights $\theta^Q$ and $\theta^\mu$\\
		Initialize target network $Q^{'}$ and $\mu^{'}$ with weights $\theta^{Q^{'}} \leftarrow  \theta^Q$, $\theta^{\mu^{'}} \leftarrow  \theta^{\mu}$\\
		
		%\KwOut{Observation $S_0$ and choose$A_0\sim\pi_\theta(S_0$)}
		
		\For{episode $ e \in \{$1,...,M$\}$}
		{Initialize a random process $\mathcal{N}$ for action exploration\\
			Receive initial observation state $s_1$\\
			\For{$t \in \{$1,...,T$\}$}
			{
				\# $S$ is the set of states in $FIFO_{\textrm{AN2N}}$\\
				\uIf {\rm{Similarity($s_t, S$)}} {
					$\mathcal{N}_t$ = $\mathcal{N}_{\textrm{big}}$
				} \Else{
					$\mathcal{N}_t$ = $\mathcal{N}_{\textrm{small}}$
				}
				Select action $a_t=\mu(s_t|\theta^\mu)+\mathcal{N}_t$ 
				according to the current policy and exploration noise\\
				Execute action $a_t$ and observe reward $r_t$ and observe new state $s_{t+1}$\\
				Store transition $(s_t, a_t, r_t, s_{t+1})$ in $R$\\
				Test the agent and store the trajectory $(s_t, r_t)$ in $R_{\textrm{AN2N}}$\\
				Calculate the cumulative discount rewards of each state:\\
				\begin{center}
					\vspace{2ex}
					$Reward(s_t) \approx r_t+\gamma r_{t+1}+\cdots+\gamma^{T-t+1}r_{T-1}+\gamma^{T-t}Q^{'}(s_T,\mu(s_T|\theta^{\mu})|\theta^{Q^{'}})$
				\end{center}
				Save the clip$({20\times(\frac{\textrm{average} \  \textrm{reward}}{\textrm{reward}}})^2, K_{\textrm{lower}}, K_{\textrm{upper}}) $\\$  Reward(s)$ minimum key states in $FIFO_{\textrm{AN2N}}$\\
				\If{t \rm{mod} u}
				{
					Sample a random minibatch of $N$ transitions $(s_i, a_i, r_i, s_{i+1})$ from $R$\\
					Set $y_i=r_i+\gamma Q^{'}(s_{i+1},\mu^{'}(s_{i+1}|\theta^{\mu^{'}})|\theta^{Q^{'}})$\\
					Update critic by minimizing the loss:$L=\frac{1}{N} \sum_{i}(y_i-Q(s_i, a_i|\theta^Q))^2$\\
					Update the actor policy using the sampled policy gradient:\\
					\begin{center}
						\vspace{1ex}
						$\nabla_{\theta^{\mu}}J\approx\frac{1}{N} \sum\limits_{i}\nabla_aQ(s, a|\theta^Q)|_{s=s_i, a=\mu(s_i)}\nabla_{\theta^{\mu}}\mu(s|\theta^{\mu})|s_i$
					\end{center}
					Update the target networks:
					\begin{center}
						$\theta^{Q^{'}}\leftarrow\tau\theta^Q+(1-\tau)\theta^{Q^{'}}$\\
						$\theta^{\mu^{'}}\leftarrow\tau\theta^{\mu}+(1-\tau)\theta^{Q^{\mu^{'}}}$
					\end{center}
				}
			}
		}
	\end{spacing}
\end{algorithm}
\begin{algorithm}[h]
	\begin{spacing}{0.8}
		\caption{SAC with AN2N}\label{algorithm2}
		\KwIn{Noise standard deviation  $\sigma_{\textrm{small}}$,  $\sigma_{\textrm{big}}$, Replay buffer size  $R$, $R_{\textrm{AN2N}}$, $FIFO_{\textrm{AN2N}}$,  Key states $K_{\textrm{upper}}$,  $K_{\textrm{lower}}$,  Temperature parameter $\alpha$}
		Randomly initialize critic networks $Q(s,a|\theta_1^Q)$, $Q(s,a|\theta_2^Q)$, and actor $\mu(s|\theta^\mu)$ with weights $\theta_1^Q$, $\theta_2^Q$ and $\theta^\mu$\\
		Initialize target network $Q^{'}$ and $\mu^{'}$ with weights $\theta_1^{Q^{'}} \leftarrow  \theta_1^Q$, $\theta_2^{Q^{'}} \leftarrow  \theta_2^Q$\\
		
		\For{episode $ e \in \{$1,...,M$\}$}
		{Initialize a random process $\mathcal{N}$ for action exploration\\
			Receive initial observation state $s_1$\\
			\For{$t \in \{$1,...,T$\}$}
			{
				\# $S$ is the set of states in $FIFO_{\textrm{AN2N}}$\\
				\uIf {\rm{Similarity($s_t, S$)}} {
					$\sigma$ = $\sigma_{\textrm{big}}$
				} \Else{
					$\sigma$ = $\sigma_{\textrm{small}}$
				}
				Select action $a_t \sim \mathcal{N}(\mu(s_t|\theta^\mu),\sigma)$ 
				according to the current policy\\
				Execute action $a_t$ and observe reward $r_t$ and observe new state $s_{t+1}$\\
				Store transition $(s_t, a_t, r_t, s_{t+1})$ in $R$\\
				Test the agent and store the trajectory $(s_t, r_t)$ in $R_{\textrm{AN2N}}$\\
				Calculate the cumulative discount rewards of each state:\\
				\begin{center}
					\vspace{2ex}
					$Reward(s_t) \approx r_t+\gamma r_{t+1}+\cdots+\gamma^{T-t+1}r_{T-1}+\gamma^{T-t}Q^{'}(s_T,\mu(s_T|\theta^{\mu})|\theta^{Q^{'}})$
				\end{center}
				Save the clip$({20\times(\frac{\textrm{average} \  \textrm{reward}}{\textrm{reward}}})^2, K_{\textrm{lower}}, K_{\textrm{upper}}) $\\$ Reward(s)$ minimum key states in $FIFO_{\textrm{AN2N}}$\\
				\If{t \rm{mod} u}
				{
					Sample a random minibatch of $N$ transitions $(s_i, a_i, r_i, s_{i+1})$ from $R$\\
					Set $y_{i}=r_i+\gamma(\min_{j=1,2} Q^{'}(s_{i+1},\mu(s_{i+1}|\theta^{\mu})|\theta_j^{Q^{'}})-\alpha$ log $\mu(s_{t+1}|\theta^\mu))$\\
					Update critic (soft Q-function) by minimizing the loss:\\
					\begin{center}
						$L=\frac{1}{N} \sum_{i}\sum_{j}(y_i-Q(s_i, a_i|\theta_j^Q))^2$
					\end{center}
					Update the actor policy using the sampled policy gradient:\\
					\begin{center}
						$\nabla_{\theta^{\mu}}J\approx\frac{1}{N} \sum\limits_{i}((\nabla_aQ(s, a|\theta^Q)|_{s=s_i, a=\mu(s_i)}\nabla_{\theta^{\mu}}\mu(s|\theta^{\mu})|s_i$ \\
						$- \nabla_a$ log $\mu(s_t|\theta^\mu))-\nabla_{\theta^{\mu}}$log $\mu(s_t|\theta^\mu)|s_i)$
					\end{center}
					Update the target networks:
					\begin{center}
						$\theta_j^{Q^{'}}\leftarrow\tau\theta_j^Q+(1-\tau)\theta_j^{Q^{'}}$
					\end{center}
				}			
			}
		}
	\end{spacing}
\end{algorithm}

\section*{ACKNOWLEDGMENT}

We would like to thank Feng Pan, Weixing Li, Xiaoxue Feng, Yan Gao, Shengyang Ge and many others at Institute of Pattern Recognition and Intelligent System of BIT for insightful discussions and valuable suggestions.

%%%%%%%%%%%%%%%%%%%%%%%%%%%%%%%%%%%%%%%%%%%%%%%%%%%%%%%%%%%%%%%%%%%%%%%%%%%%%%%%


\clearpage
\begin{thebibliography}{99}
\bibitem{c1} Richard S Sutton and Andrew G Barto. Reinforcement learning: An introduction.  MIT press, 2018.
\bibitem{c2} Fortunato, Meire, Mohammad Gheshlaghi Azar, Bilal Piot, Jacob Menick, Matteo Hessel, Ian Osband, Alex Graves et al. "Noisy Networks For Exploration." In International Conference on Learning Representations. 2018.
\bibitem{c3} Williams R J. Simple statistical gradient-following algorithms for connectionist reinforcement learning[J]. Machine learning, 1992, 8(3): 229-256.
\bibitem{c4} Haarnoja, Tuomas, Aurick Zhou, Pieter Abbeel, and Sergey Levine. "Soft actor-critic: Off-policy maximum entropy deep reinforcement learning with a stochastic actor." In International conference on machine learning, pp. 1861-1870. PMLR, 2018.
\bibitem{c5} Schmidhuber, Jurgen. "A possibility for implementing curiosity and boredom in model-building neural controllers." In Proc. of the international conference on simulation of adaptive behavior: From animals to animats, pp. 222-227. 1991.
\bibitem{c6} Oudeyer, Pierre-Yves, Frdric Kaplan, and Verena V. Hafner. "Intrinsic motivation systems for autonomous mental development." IEEE transactions on evolutionary computation 11, no. 2 (2007): 265-286.
\bibitem{c7} Haarnoja, Tuomas, Haoran Tang, Pieter Abbeel, and Sergey Levine. "Reinforcement learning with deep energy-based policies." In International Conference on Machine Learning, pp. 1352-1361. PMLR, 2017.
\bibitem{c8} Osband, Ian, Benjamin Van Roy, and Zheng Wen. "Generalization and exploration via randomized value functions." In International Conference on Machine Learning, pp. 2377-2386. PMLR, 2016.
\bibitem{c9} Osband, Ian, Charles Blundell, Alexander Pritzel, and Benjamin Van Roy. "Deep exploration via bootstrapped DQN." Advances in neural information processing systems 29 (2016): 4026-4034.
\bibitem{c10} Touati, Ahmed, Harsh Satija, Joshua Romoff, Joelle Pineau, and Pascal Vincent. "Randomized value functions via multiplicative normalizing flows." In Uncertainty in Artificial Intelligence, pp. 422-432. PMLR, 2020.
\bibitem{c11} Auer, Peter. "Using confidence bounds for exploitation-exploration trade-offs." Journal of Machine Learning Research 3, no. Nov (2002): 397-422.
\bibitem{c12} Bellemare, Marc, Sriram Srinivasan, Georg Ostrovski, Tom Schaul, David Saxton, and Remi Munos. "Unifying count-based exploration and intrinsic motivation." Advances in neural information processing systems 29 (2016): 1471-1479.
\bibitem{c13} Ostrovski, Georg, Marc G. Bellemare, Aaron Oord, and Remi Munos. "Count-based exploration with neural density models." In International conference on machine learning, pp. 2721-2730. PMLR, 2017.
\bibitem{c14} Zhao, Rui, and Volker Tresp. "Curiosity-driven experience prioritization via density estimation." arXiv preprint arXiv:1902.08039 (2019).
\bibitem{c15} Tang, Haoran, Rein Houthooft, Davis Foote, Adam Stooke, Xi Chen, Yan Duan, John Schulman, Filip De Turck, and Pieter Abbeel. " Exploration: A Study of Count-Based Exploration for Deep Reinforcement Learning." In NIPS. 2017.
\bibitem{c16} Martin, Jarryd, S. Suraj Narayanan, Tom Everitt, and Marcus Hutter. "Count-based exploration in feature space for reinforcement learning." In Proceedings of the 26th International Joint Conference on Artificial Intelligence, pp. 2471-2478. 2017.
\bibitem{c17} Houthooft, Rein, Xi Chen, Yan Duan, John Schulman, Filip De Turck, and Pieter Abbeel. "VIME: Variational Information Maximizing Exploration." Advances in Neural Information Processing Systems 29 (2016): 1109-1117.
\bibitem{c18} Pathak, Deepak, Pulkit Agrawal, Alexei A. Efros, and Trevor Darrell. "Curiosity-driven exploration by self-supervised prediction." In International conference on machine learning, pp. 2778-2787. PMLR, 2017.
\bibitem{c19} Burda, Yuri, Harri Edwards, Deepak Pathak, Amos Storkey, Trevor Darrell, and Alexei A. Efros. "Large-Scale Study of Curiosity-Driven Learning." In International Conference on Learning Representations. 2018.
\bibitem{c20} Stadie, Bradly C., Sergey Levine, and Pieter Abbeel. "Incentivizing exploration in reinforcement learning with deep predictive models." arXiv preprint arXiv:1507.00814 (2015).
\bibitem{c21} Pathak, Deepak, Dhiraj Gandhi, and Abhinav Gupta. "Self-supervised exploration via disagreement." In International conference on machine learning, pp. 5062-5071. PMLR, 2019.
\bibitem{c22} Guo, Youtian, Qi Gao, and Feng Pan. "Trained Model Reuse of Autonomous-Driving in Pygame with Deep Reinforcement Learning." In 2020 39th Chinese Control Conference (CCC), pp. 5660-5664. IEEE, 2020.
\bibitem{c23} De Baar, H. J. W. "von Liebig's law of the minimum and plankton ecology (1899?1991)." Progress in oceanography 33, no. 4 (1994): 347-386.
\bibitem{c24} Bellman, Richard, and Robert E. Kalaba. Dynamic programming and modern control theory. Vol. 81. New York: Academic Press, 1965.
\bibitem{c25} Bertsekas, Dimitri P. Dynamic programming and optimal control: Vol. 1. Belmont: Athena scientific, 2000.
\bibitem{c26} Watkins, Christopher JCH, and Peter Dayan. "Q-learning." Machine learning 8, no. 3-4 (1992): 279-292.
\bibitem{c28} Mnih, Volodymyr, Koray Kavukcuoglu, David Silver, Andrei A. Rusu, Joel Veness, Marc G. Bellemare, Alex Graves et al. "Human-level control through deep reinforcement learning." nature 518, no. 7540 (2015): 529-533.
\bibitem{c29} Lillicrap, Timothy P., Jonathan J. Hunt, Alexander Pritzel, Nicolas Heess, Tom Erez, Yuval Tassa, David Silver, and Daan Wierstra. "Continuous control with deep reinforcement learning." In ICLR (Poster). 2016.
\bibitem{c30} Schulman, John, Sergey Levine, Pieter Abbeel, Michael Jordan, and Philipp Moritz. "Trust region policy optimization." In International conference on machine learning, pp. 1889-1897. PMLR, 2015.
\bibitem{c31} Mnih, Volodymyr, Adria Puigdomenech Badia, Mehdi Mirza, Alex Graves, Timothy Lillicrap, Tim Harley, David Silver, and Koray Kavukcuoglu. "Asynchronous methods for deep reinforcement learning." In International conference on machine learning, pp. 1928-1937. PMLR, 2016.
\bibitem{c32} Fujimoto, Scott, Herke Hoof, and David Meger. "Addressing function approximation error in actor-critic methods." In International Conference on Machine Learning, pp. 1587-1596. PMLR, 2018.
\bibitem{c33} Peng, Jing, and Ronald J. Williams. "Incremental multi-step Q-learning." In Machine Learning Proceedings 1994, pp. 226-232. Morgan Kaufmann, 1994.
\bibitem{c34} Todorov, Emanuel, Tom Erez, and Yuval Tassa. "Mujoco: A physics engine for model-based control." In 2012 IEEE/RSJ International Conference on Intelligent Robots and Systems, pp. 5026-5033. IEEE, 2012.







\end{thebibliography}
\end{document}